\documentclass{article}
\usepackage{ijcai20}

\usepackage{times}
\usepackage{soul}
\usepackage{url}
\usepackage[hidelinks]{hyperref}
\usepackage[utf8]{inputenc}
\usepackage[small]{caption}
\usepackage{graphicx}
\usepackage{amsmath}
\usepackage{amssymb}
\usepackage{amsthm}
\usepackage{booktabs}
\usepackage{algorithm}
\usepackage{algorithmic}
\usepackage{xcolor}
\title{Batch Decorrelation for Active Metric Learning}

\author{
Priyadarshini Kumari $^{1}$
\and
Ritesh Goru$^{1}$\and
Siddhartha Chaudhuri$^{1,2}$\And
Subhasis Chaudhuri$^{1}$
\affiliations
$^1$IIT Bombay\\
$^2$Adobe Research
\emails
priyadarshini.k@iitb.ac.in,
riteshgoru@iitb.ac.in,
sidch@cse.iitb.ac.in,
sc@ee.iitb.ac.in
}

\begin{document}

\maketitle

\begin{abstract}
We present an active learning strategy for training parametric models of distance metrics, given triplet-based similarity assessments: object $x_i$ is more similar to object $x_j$ than to $x_k$. In contrast to prior work on class-based learning, where the fundamental goal is classification and any implicit or explicit metric is binary, we focus on {\em perceptual} metrics that express the {\em degree} of (dis)similarity between objects. We find that standard active learning approaches degrade when annotations are requested for {\em batches} of triplets at a time: our studies suggest that correlation among triplets is responsible. In this work, we propose a novel method to {\em decorrelate} batches of triplets, that jointly balances informativeness and diversity while decoupling the choice of heuristic for each criterion. Experiments indicate our method is general, adaptable, and outperforms the state-of-the-art.
\end{abstract}

\section{Introduction}

Defining a good distance metric over a space of objects is important for many machine learning tasks, such as classification, clustering and object segmentation. Often, such a metric represents binary relationships among objects: for instance, it may consider two objects similar if they belong to the same class and dissimilar otherwise. However, domains like cognitive science and human-computer interaction (HCI) often require capturing the more granular and complex notion of {\em perceptual} distance: a {\em continuous} measure of the perceived dissimilarity between two objects. This allows {\em relative} assessments such as: object $x$ is more similar to $y$ than to $z$, even when all three objects are from the same class. Indeed, it is even possible for two objects from different classes (say, a dolphin and a fish) to be perceptually more similar than two objects from the same class (the dolphin and any land mammal). Learning a robust predictor of the perceptual distance between any two objects from some input domain, given training data over a subset of that domain, is therefore an important and challenging problem. In recent research, this predictor is commonly a deep neural network.

While regressing a metric from numerical measurements of the distances between pairs of objects is straightforward, it is usually very difficult for human annotators to accurately and consistently estimate such distances. 
Instead, it is easier to answer {\em ranking} questions: is object $x$ more similar to object $y$ or $z$? Such ``triplet'' comparisons are the backbone of metric learning, in both binary (categorization) and continuous (perceptual) contexts. The number of possible triplet comparisons grows cubically -- $O(n^3)$ for $n$ objects -- and the cost of annotating them all is prohibitive. However, for reasonably well-behaved metrics the triplets contain much redundancy, since the underlying measure is assumed to at least approximately follow the triangle inequality. This suggests that good models can be trained on much smaller triplet sets.

\begin{figure}[!t]
\centering
\vspace{-5mm}
\begin{tabular}{ccc}
\includegraphics[scale=0.4]{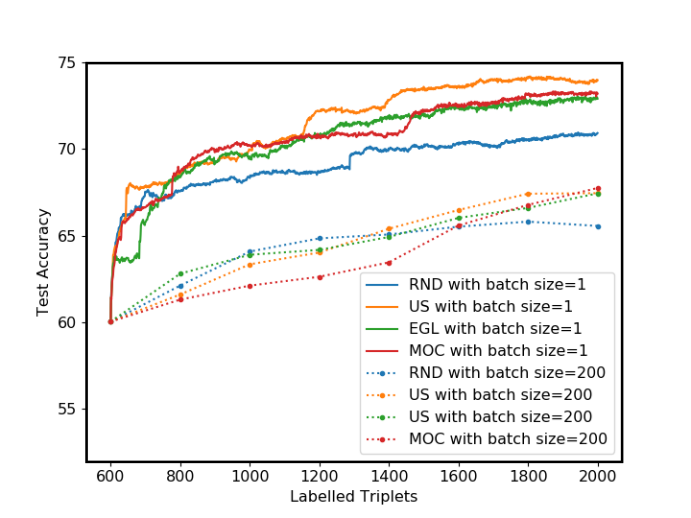}
\end{tabular}
\vspace{-4mm}
\caption{Negative impact of correlated triplets on active learning. Deep metric learning with 3 standard AL heuristics (uncertainty (US), expected gradient length (EGL), model output change (MOC)) all outperform random sampling (RND) when triplets are added one at a time from the Yummly food dataset. With 200-triplet batches, they are comparable to or worse than random. (Note: batch size also affects performance with random sampling because we mimic real-world computational budgets by allotting a fixed (high) number of training epochs after each batch. When training with single triplets, the network gets to train more often, hence longer overall.)}
\vspace{-1mm}
\label{incr1}
\end{figure}

The main challenge, then, is selecting {\em which} small subset of triplets should be annotated. This task is addressed by {\em active learning}, a standard technique to minimize effort by iteratively requesting annotations for the most informative samples (here, triplets) from a large pool of unlabeled data. In each iteration, a batch of samples is labeled and added to the training set, and the model is updated. The process repeats until a desired accuracy is achieved. Active learning has been effective for many applications \cite{adomavicius2005toward,calinon2007learning,settles2012,yang2017suggestive,zhang2017active}, but only a few studies \cite{tamuz2011adaptively,heim2015active,lohaus2019uncertainty} employ it for triplet-based perceptual metric learning. These methods are non-parametric (the majority use multidimensional scaling (MDS)) and cannot predict the distance between novel objects at test time. But they also have another significant shortcoming. To reduce the incremental training cost and enable efficient use of human resources, annotations are typically requested in {\em batches} of, say, a few hundred triplets at a time, after which the model is updated taking the new training batch into account. Prior active learning strategies often perform worse than baseline random sampling when operating in batch mode (Figure \ref{incr1}). We find that {\em correlation} among samples in a batch is responsible: individually informative samples are likely to form strongly correlated clusters, lowering the expected benefit. There have been very recent efforts to alleviate this for classification~\cite{kirsch2019batchbald,ash2019deep}, but none for metric learning.

In this paper, we propose the {\bf first batch decorrelation strategy developed specifically for triplet-based active metric learning}. Our method jointly balances informativeness and diversity of a batch of triplets while decoupling the choice of heuristic for each criterion (the approach of Ash et al.~\shortcite{ash2019deep}, developed for classification, represents both using the same unnormalized gradients). We select an overcomplete set of informative triplets, and find well-separated yet informative representatives from them using farthest-point sampling, a technique borrowed from visual computing~\cite{eldar1997farthest}. We develop several alternative measures of triplet separation, and compare their effectiveness over a range of different input modalities, tasks, and hyperparameter settings. Our framework is robust and outperforms the current state-of-the-art in all evaluated scenarios. We demonstrate the method for neural network-based learners, but it can be easily adapted to any parametric learning scenario.

\section{Related Work}
\label{sec:related}

We overview related work on active learning in three categories. For a general survey of deep metric learning, we recommend recent papers such as~\cite{priyadarshini2019perceptnet}.

\subsection{Active learning for class-based learning.}
There is significant work on active learning for \emph{classification} \cite{settles2012}, exploring heuristics like uncertainty sampling, query-by-committee, expected model change, expected error/variance minimization, and information gain \cite{settles2012,krishnamurthy2017active,bachman2017learning,konyushkova2017learning}. Among these, uncertainty sampling is an effective and widely used strategy. Two other interesting works \cite{hoi2006batch,wei2015submodularity} select a batch of diverse and informative samples by posing active learning as a submodular subset selection problem. In contrast to our approach, all these methods focus on object classification.

\subsection{Active learning for deep learning.}
Several recent works apply active learning heuristics to neural networks. Sener and Savarese \shortcite{CNN2018ICLR} define active learning as a core-set problem to select diverse samples. Sinha et al. \shortcite{sinha2019variational} adversarially select outliers to the distribution of labeled samples. Gal et al. \shortcite{gal2017deep} propose Bayesian active learning (BALD) for image classification, maximizing mutual information between model parameters and model prediction. It works well when selecting samples one at a time but fails for larger batches. To address this, Kirsch et al. \shortcite{kirsch2019batchbald} introduce a modified BALD for batches. They optimize informativeness and diversity by maximizing mutual information between selected points and model parameters.

Ash et al.~\shortcite{ash2019deep} present another method (BADGE) for diversification where samples with diverse gradient vectors are picked for annotation. However, they use the gradient w.r.t. the most probable label instead of the expected gradient, and select a batch using $k$-means++ \cite{arthur2006k} over all unlabeled samples in the gradient space. This has shortcomings: (1) the gradient is used for {\em both} informativeness and diversity, whereas our method decouples these choices to leverage the best individual heuristics; (2) $k$-means++ will pick sample centroids around the origin (small gradient vectors) if there are dense clusters there, hence the bias towards more informative samples is reduced; and (3) $k$-means++ requires an explicit triplet embedding and cannot work with just a distance measure, unlike our farthest-point sampling. Experiments indicate that our approach consistently outperforms BADGE adapted to metric learning.

All these methods share our goal of efficient learning from fewer annotations in a batch setting, but they focus on classification problems. In contrast, we aim to learn a (perceptual, parametric) distance metric from triplet constraints. This requires {\em triplet-specific} informativeness and diversity measures, that can also be used to compare unseen data samples.

\subsection{Active triplet-based metric learning.} There are only a few papers on active metric learning from triplet-based relative constraints. Tamuz et al.~\shortcite{tamuz2011adaptively} sample triplets which reduce the uncertainty in object location in the embedding space. Heim et al.~\shortcite{heim2015active} propose a similar approach and show that perceptual similarity can be modeled with even fewer triplets if auxiliary features are used. Again, both methods work well in single-triplet mode but fail with neural networks in a batch setting. Moreover, both methods are non-parametric and designed to achieve data efficiency for MDS: they cannot predict the distance between novel objects. Recently, Lohaus et al.~\shortcite{lohaus2019uncertainty} proposed a triplet uncertainty estimate by modeling triplet ordering as a Gaussian process, a natural choice for active learning as it measures prediction confidence. However, similar to MDS-based approaches, it is non-parametric and the learning complexity is $O(n^{3})$. Finally, none of these methods optimize diversity in addition to informativeness during batch selection.

\section{Method}
\label{sec:method}

In this section, we describe our framework for batch-mode active metric learning. We first briefly define the problem of perceptual metric learning from triplet comparisons, using a deep network as the regressor. Next, we discuss an active data sampling approach for learning the metric efficiently. Finally, as our main technical contribution, we introduce a method for decorrelating batches of triplets, to address the known limitations of active learning with large incremental batches.

\subsection{Triplet-Based Deep Metric Learning}

We are given a set of objects \mbox{$X = \{x_i\}_1^n \in \mathbb{R}^d$}, described by their $d$-dimensional features. We also have relative comparisons of their pairwise distances as a set of triplets \mbox{$C = \{(x_i, x_j, x_k)\}$}. Here, each triplet indicates that the anchor object $x_i$ is more similar to $x_j$ than to $x_k$. Our goal is to learn a distance metric \mbox{$d : \mathbb{R}^d \times \mathbb{R}^d \rightarrow \mathbb{R}$} which preserves the triplet orderings -- \mbox{$d(x_i, x_j) < d(x_i, x_k)$}. As is standard, we assume the comparisons can be noisy and inconsistent, hence the solution is approximate and we only try to maximize the number of preserved orderings. Although this framework applies to any kind of metric learning, including binary metrics for classification, our focus is to learn perceptual metrics reflecting continuously varying degrees of (dis)similarity.

Typically, perceptual (dis)similarity between objects is modeled by projecting the data using a learned embedding kernel \mbox{$\phi:\mathbb{R}^d\rightarrow \mathbb{R}^{\hat{d}}$} such that the distance between two objects \mbox{$d_{\phi}(x,y) = \|\phi(x) - \phi(y)\|$} reflects the perceptual distance. Following prior work~\cite{lu2017matchnet,perceptualMetric,priyadarshini2019perceptnet}, we use a deep network to represent the embedding kernel $\phi$. For each triplet, we pass its three objects through three copies of the network with shared weights. We train the network using an exponential triplet loss that tries to maximize the squared distance margin \mbox{$d^2_\phi(x_i, x_k) - d^2_\phi(x_i, x_j)$} for each triplet \mbox{$(x_i, x_j, x_k)$}:
\begin{equation} \label{loss}
\mathcal{L}(x_i, x_j, x_k, \phi) = \sum_{C} e^{ -(d^2_\phi(x_i, x_k) - d^2_\phi(x_i, x_j)) }
\end{equation}
The model is trained using standard back-propagation. In all our experiments, the network comprises a series of fully connected layers with ReLU nonlinearities. The hyperparameters are manually tuned for each dataset, as detailed in Section \ref{sec:results}.

\subsection{Deep Active Metric Learning}

The number of possible triplet comparisons of $n$ objects grows cubically -- $O(n^3)$ -- and gathering annotations for all of them is prohibitively expensive. However, for a reasonably smooth metric that at least approximately respects triangle inequality, a much smaller subset of ``informative'' triplets is usually sufficient to capture its salient features. Active learning is a standard technique to discover such a subset. It is an iterative method that greedily repeats the following two steps. First, given an incremental labeling budget $b$, and the set of previously labeled triplets $L$, it selects a batch $T$ of $b$ new triplets (from some potentially large set of unlabeled triplets $U$) and requests annotations (i.e. ground-truth orderings) for them. Second, the batch is added to the set of labeled triplets and the metric model is retrained. The process repeats till some desired accuracy is reached. The main technical challenge is to choose the most informative new triplets in each batch, such that the model reaches its asymptotic accuracy in as few iterations as possible.

The informativeness of a triplet can be measured by various heuristics, such as model uncertainty, prediction variance, or mutual information. In this paper, we adopt model uncertainty as a representative measure which has proved to be very effective in practice~\cite{settles2012}, though we also show experiments with other measures. We consider a triplet ``uncertain'', and hence informative, if the model is highly uncertain in predicting its ordering: $d_{\phi}(x_i, x_k) \approx d_{\phi}(x_i, x_j)$. A probability distribution over the triplet response is a natural way to measure uncertainty~\cite{tamuz2011adaptively}:
\vspace{-1mm}
\begin{equation} \label{dist}
p_{ijk} = p(t|\phi) = \frac{\mu + d_\phi^2(x_i, x_k)}{2\mu + d_\phi^2(x_i, x_k) + d_\phi^2(x_i, x_j)}
\vspace{-0.5mm}
\end{equation}
Here, $p_{ijk}$ denotes the probability of object $i$ being closer to $j$ than to $k$ in the embedding space, $\mu >= 0$ is a hyperparameter, and $p_{ijk}+p_{ikj}=1$.
The model's uncertainty about the unlabeled triplet can then be expressed as the expected Shannon entropy over all possible orderings: \mbox{$f_{\phi}(t) = -p_{ijk} \log p_{ijk} - p_{ikj}\log p_{ikj}$}. The score of a selected batch $T$, na\"{i}vely ignoring correlations between triplets, can be defined by summing up the individual uncertainties:
\vspace{-0.5mm}
\begin{equation} \label{entropy}
f_{\phi}(T) \ = \ \sum_{t \in T} f_{\phi}(t)
\vspace{-1mm}
\end{equation}
However, simply using this score to select each incremental batch, as \mbox{$T^* = \arg \max_{T \subset U, |T| = b} f_{\phi}(T)$}, is fast (it reduces to picking the $b$ top-scoring triplets) but suboptimal. The top $b$ triplets are individually informative, but may not be so jointly. For instance, they may all be tightly clustered in the single most uncertain region in the embedding space, having high redundancy and giving the model no wider information. To address this shortcoming, we choose a {\em diverse yet informative} batch by applying decorrelation measures, discussed next.

\subsection{Triplet Batch Decorrelation}
\label{sec:decorr}

Selecting a single triplet in each iteration ($b = 1$) is highly inefficient since it requires infeasibly frequent retraining. To reduce this overhead, the standard approach is to pick larger batches ($b \gg 1$), but developing an optimal batch selection policy is difficult because of correlation between the most informative triplets as discussed above (in other words, informativeness is not compositional). In the general case, the problem is NP-hard. Some prior works develop effective greedy approximations by defining batch utility as a monotone submodular function~\cite{golovin2011adaptive}, but making the utility function submodular is not always natural.

Instead, we propose a two-step approach, each step involving greedy sampling. The first step emphasizes {\bf informativeness}: we pick an {\em overcomplete} set $S$ of the most informative individual triplets, of much larger size than $b$, from the unlabeled set $U$. The second step emphasizes {\bf diversity}: we {\em subsample} $S$ to select a final batch $T^*$ that effectively covers $S$, given a measure $\rho_{\phi}$ of the distance (non-correlation) between triplets conditioned on the current model $\phi$. Picking the batch that minimizes the distance of each triplet in $S$ to its nearest triplet in $T^*$ is also NP-hard. However, it has been shown that farthest-point-sampling (FPS), a greedy approach originally developed for image processing~\cite{eldar1997farthest}, gives a near-optimal solution. In our adaptation of FPS, we begin with the most separated pair of triplets, and then recursively select the unselected triplet that is farthest from the selected set. Formally, the recursion is defined as \mbox{$T \leftarrow T \cup \arg \max_{t \in S \setminus T} \min_{t' \in T} \rho_{\phi}(t, t')$}. Combining the two steps, as formalized in Algorithm \ref{alg:algorithm}, we obtain a diverse yet informative batch.

\begin{algorithm}[t!]
\caption{Batch-Mode Active Learning}
\label{alg:algorithm}
\textbf{Input}: Neural network architecture $\phi$, object features $X \subset \mathbb{R}^d$, labeled triplets $L$, unlabeled triplets $U$, initial pool size $l$, batch size $b$, oversampling size $k > b$ \\
\textbf{Initialize}: Train initial model $\phi_{0}$ on $l$ randomly selected labeled triplets $L_{0}$.
\begin{algorithmic}[1] 
\FOR{m=1,2,\dots,M}
\STATE $S_m = \arg \max_{S \subset U, |S| = k} f_{\phi}(S)$ using Eq. \ref{entropy}
\STATE $T_m^* = \arg \max_{\{t_i, t_j\} \subset S_m} \rho_{\phi}(t_i, t_j)$
\FOR{n=3,\dots,b}
\STATE $T_m^* \leftarrow T_m^* \cup \{ \arg \max_{t \in S_m \setminus T_m^*} \min_{t' \in T_m^*} \rho_{\phi}(t, t') \}$
\ENDFOR
\STATE $L \leftarrow L \cup T_m^*$
\STATE Train starting from model $\phi_{m - 1}$, on labeled triplets $L$ with loss in Eq. \ref{loss}, to obtain updated model $\phi_m$.
\ENDFOR
\STATE \textbf{return} Final model $\phi_M$
\end{algorithmic}
\end{algorithm}

The remaining challenge is to define $\rho_{\phi}$. To incorporate triplet informativeness in this measure as well, in case $S$ is not uniformly informative, we define $\rho_{\phi}(t, t') := f_{\phi}(t) \times f_{\phi}(t') \times \gamma_{\phi}(t, t')$, discouraging uninformative triplets from being selected even if they are far apart. Now, we propose several intuitive new ways of defining the purely ``geometric'' term $\gamma_{\phi}(t, t')$, since sample-wise (dis)similarity measures like $\textrm{L}_p$ or cosine metrics do not directly apply to unordered triplets.

\begin{enumerate}
    \item {\bf Gradient distance:} Intuitively, two triplets can be considered dissimilar if they change the model differently. Since the neural network is optimized using gradient descent, we represent each triplet by the expected gradient (since the ordering is unknown) of its loss w.r.t. the last layer $\phi_{\textrm{out}}$ of the network:
    \vspace{-0.5mm}
    \begin{equation} \label{exp_grad}
    g(t)=p_{ijk} \frac{\partial \mathcal{L}(x_i, x_j, x_k, \phi)}{\partial \phi_{\textrm{out}}} + p_{ikj} \frac{\partial \mathcal{L}(x_i, x_k, x_j, \phi)}{\partial \phi_{\textrm{out}}}
    \end{equation}
    The distance is then \mbox{$\gamma_{\phi}(t, t') = 1 - \left \langle \frac{g(t)}{|g(t)|}, \frac{g(t')}{|g(t')|} \right \rangle$}. Note the contrast with the related gradient-based decorrelation of Ash et al.~\shortcite{ash2019deep}, as detailed in Section \ref{sec:related}.

    \item {\bf Euclidean distance:} We concatenate the object embeddings of a triplet into a 3-dimensional vector: \mbox{$\phi(x_i, x_j, x_k) = \phi(x_i) \oplus \phi(x_j) \oplus \phi(x_k)$}. The expected L2 distance over possible orderings gives:
    \vspace{-1mm}
    \begin{equation}
    \gamma_{\phi}(t, t') = \sum_{ y \in \{ijk, ikj\} } \frac{1}{2} \| \phi(t^y) - \phi(t') \|,
    \end{equation}
    where $t^y$ denotes one of the two orderings of the triplet.

    \item {\bf Centroidal distance:} Instead of representing a triplet by its concatenated object embeddings, we represent it by the (ordering-independent) centroid of the embeddings of its three objects: $c(t) = \frac{1}{3}(\phi(x_i)+\phi(x_j)+\phi(x_k))$. Then, the distance between triplets is simply the L2 distance between their centroids: $\gamma_{\phi}(t, t') = \| c(t)- c(t') \|$. Note that unlike the preceding methods, this is agnostic to triplet annotation and robust to model bias.

    \item {\bf Oriented distance:} The centroid of a triplet is a single point, throwing away information about the {\em orientation} of the triplet in the embedding space. We incorporate such information into the oriented distance, defined as the sum of the distances between the anchors of two triplets, and between their ``orientation vectors'' \mbox{$\hat{r}(x_i, x_j, x_k) = \textrm{unit}\left( \phi(x_k) + \phi(x_j) - 2\phi(x_i) \right)$}:
    \vspace{-1mm}
    \begin{equation}
    \gamma_{\phi}(t, t') = d_{\phi}(x_i, x'_i) + \left( 1 - \langle \hat{r}(t), \hat{r}(t') \rangle \right),
    \end{equation}
    \vspace{-1mm}
    where $x_i, x'_i$ are the anchor objects of $t, t'$ respectively.
\end{enumerate}

\begin{figure*}[!t]
\centering
\begin{tabular}{ccc}
\includegraphics[trim=0 0 0 1mm,clip,scale=0.34]{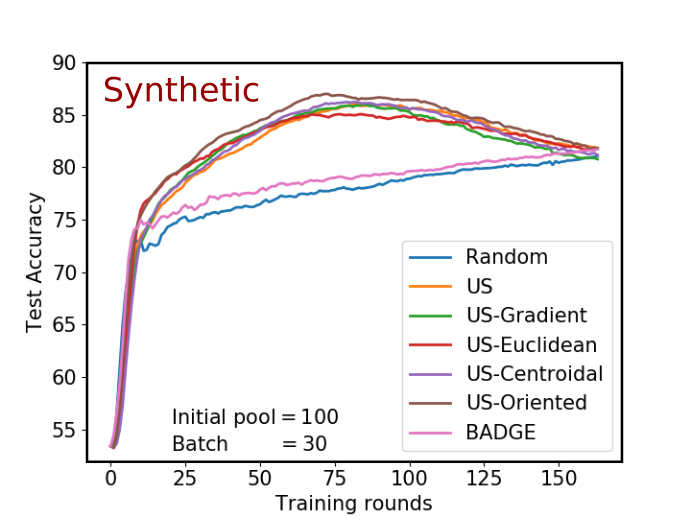} 
\vspace{-2.0mm}\hspace{-7.0mm}
\includegraphics[trim=0 0 0 1mm,clip,scale=0.34]{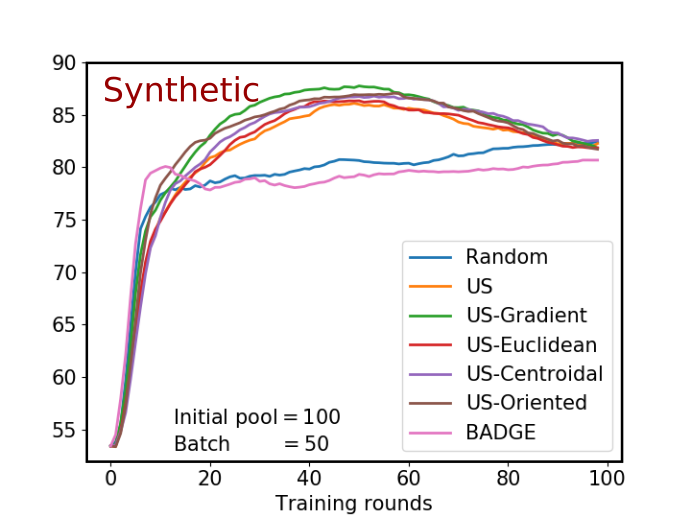}
\vspace{-2.0mm} \hspace{-7.0mm}
\includegraphics[trim=0 0 0 1mm,clip,scale=0.34]{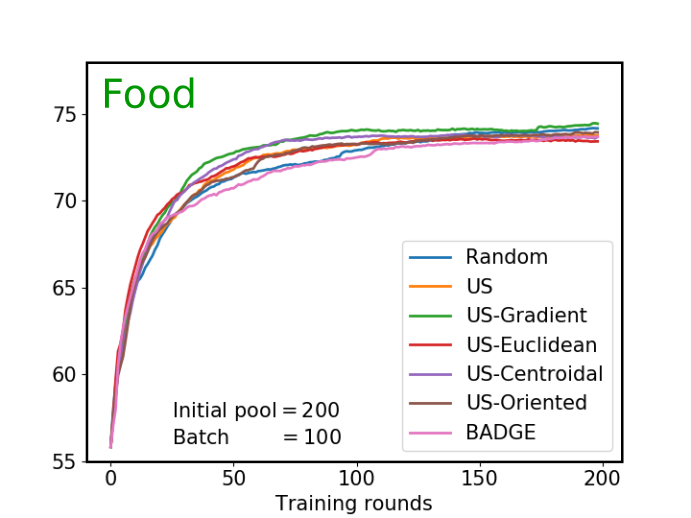}\\
\vspace{-1.1mm}
\includegraphics[trim=0 0 0 8mm,clip,scale=0.34]{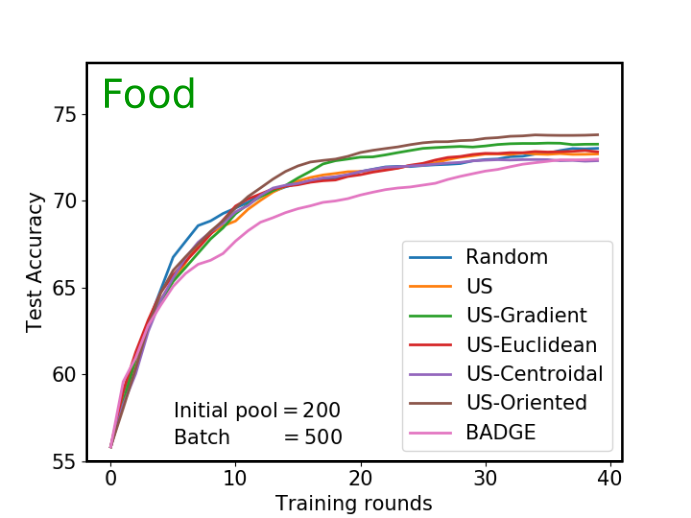}
\vspace{-1.1mm} \hspace{-7.0mm}
\includegraphics[trim=0 0 0 8mm,clip,scale=0.34]{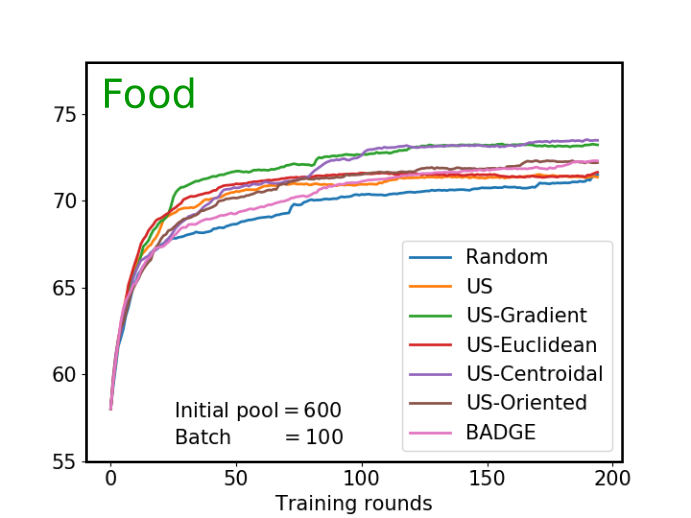}
\vspace{-1.1mm} \hspace{-7.0mm}
\includegraphics[trim=0 0 0 8mm,clip,scale=0.34]{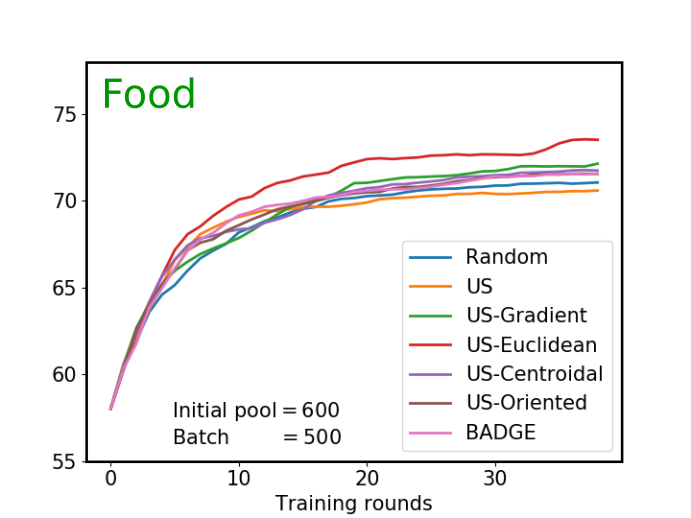}\\
\vspace{-1.0mm}
\includegraphics[trim=0 0 0 8mm,clip,scale=0.34]{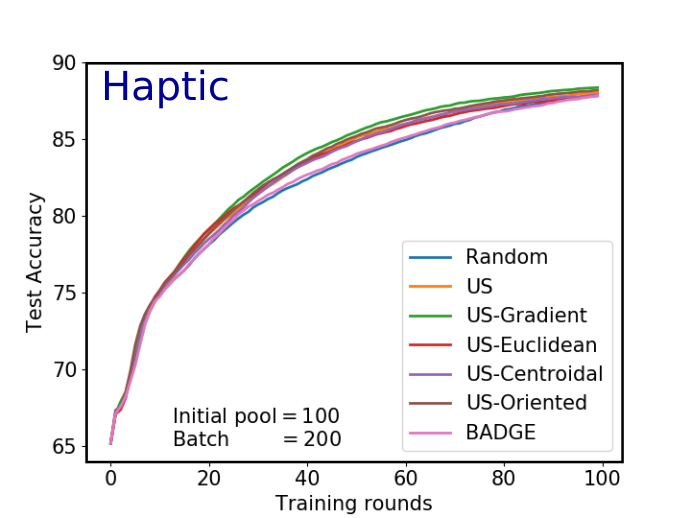}
\vspace{-1.0mm} \hspace{-7.mm}
\includegraphics[trim=0 0 0 8mm,clip,scale=0.34]{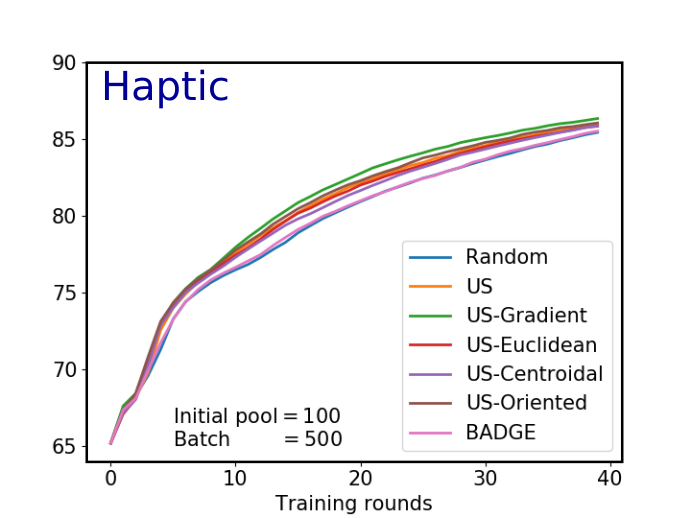}
\vspace{-1.0mm} \hspace{-7.mm}
\includegraphics[trim=0 0 0 8mm,clip,scale=0.34]{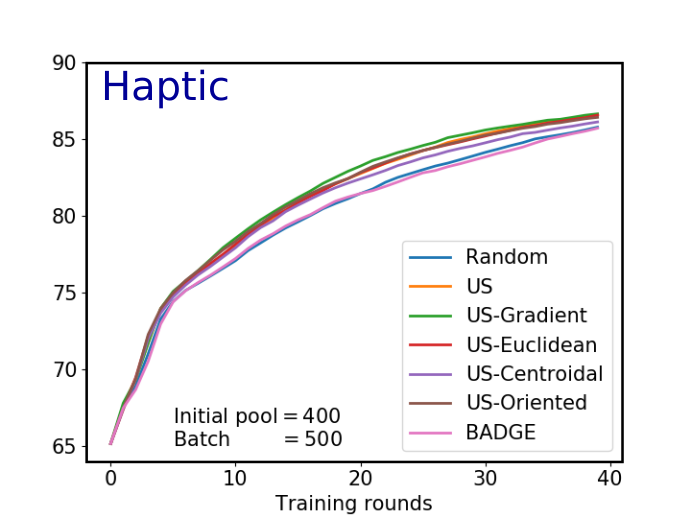}\\
\end{tabular}
\caption{Performance of active learning for different batch and initial pool sizes, on real and synthetic datasets. Different variants of our batch decorrelation method (US-{\em $\langle$Dist$\rangle$}) are evaluated and compared to three baselines: Random sampling (Random), basic uncertainty sampling without decorrelation (US), and adapted BADGE [Ash et al., 2020]. Our method consistently outperforms the baselines across all three datasets. The standard deviation ranges across datasets are: Synthetic: min $0.7\%$ (US-Gradient), max $2.55\%$ (BADGE); Food: min $1.5\%$ (US-Gradient), max $3.4\%$ (US-Gradient); Haptic: min $0.266\%$ (US-Euclidean), max $0.673\%$ (Random).}
\vspace{-3.0mm}
\label{Q1}
\end{figure*}

\section{Experiments}
\label{sec:results}

To evaluate the performance of our batch decorrelation method, we perform several experiments, addressing three questions: (Q1) How effective is our method for different batch (and initial pool) sizes, varying levels of noise rates in the training set, and different triplet diversity measures, across different datasets and modalities? (Q2) How important is decoupling the choices of informativeness and diversity heuristics, and how effective is decorrelation with different triplet informativeness measures? (Q3) How effective are individual components of our framework? We start by introducing our datasets and evaluation setup before discussing each question in subsequent sections.

\subsection{Datasets and Evaluation Setup}
We evaluate our method on two challenging real-world datasets, as well as on synthetic data.

\paragraph{Synthetic Data.} We generate 100 samples $X$ from a 10-D standard normal distribution. From $X$, we randomly sample 20K training and 20K test triplets, each with a ground truth ordering from a predefined (random) Mahalanobis metric. To check if our framework is robust to noisy data, we randomly flip the order of $20\%$ of the training triplets.

\paragraph{Yummly Food Data.} This dataset has 72148 triplets defined over 73 images of food items \cite{wilber2014cost}. The triplet constraints are based on taste similarity. Each item is represented by an L1-normalized 6-D feature vector indicating the proportions of different taste properties: salty, savory, sour, bitter, sweet, and spicy. To generate a train/test split we randomly subsample 20K training and 20K test triplets.

\paragraph{Haptic Texture Data.} This dataset has 108 surface materials (metals, paper, fabrics, etc), each represented by 10 texture signals \cite{Strese} produced by tracing a haptic stylus over the physical surface. The triplets are generated from the ground truth perceptual dissimilarity metric gathered in user experiments \cite{priyadarshini2019perceptnet}. As before, we randomly subsample 20K training and 20K test triplets. Each texture signal is represented by a 32-D CQFB spectral feature vector \cite{priyadarshini2019perceptnet}.

\paragraph{Evaluation setup.} The performance of the learned perceptual metric is evaluated after each training round, consisting of selecting one new batch of triplets followed by updating the model. We measure model perfomance by the \textit{triplet generalization accuracy} (TGA): the fraction of test triplets whose ordering is consistent with the learned metric. On each dataset, we repeat the experiment over 5 random train/test splits and report the mean accuracy after each round.

\paragraph{Variants and Baselines.} We report the performance of each of the 4 triplet distance measures in Section \ref{sec:decorr} as \mbox{\bf US-{\em $\langle$Dist$\rangle$}}, denoting the variant of our method where we first select the most informative triplets via uncertainty sampling, and then pick a diverse subset using triplet distance measure {\em $\langle$Dist$\rangle$} (e.g. gradient distance). We compare to three baselines: (1) {\bf Random}: Triplets are uniformly sampled at random; (2) {\bf US}: Uncertainty-based approach {\em without} decorrelation, which selects the top $b$ triplets with highest entropy as per Eq. \ref{entropy} \cite{tamuz2011adaptively,heim2015active}; (3) {\bf BADGE}: Gradient-based approach that selects a set of triplets with diverse loss gradients using $k$-means (the gradient for a triplet is computed for the most probable label, as opposed to our expected gradient) \cite{ash2019deep}.

\paragraph{Hyperparameters.} We manually tuned network hyperparameters in each dataset: {\em Synthetic:} 3 FC layers with 10, 20, 10 neurons resp.; {\em Food:} 3 FC layers with 6, 12, 12 neurons resp.; {\em Haptic:} 4 FC layers with 32, 32, 64, 32 neurons resp. All layers have ReLU activation. We train using Adam~\cite{kingma2014adam} with learning rate $10^{-4}$. Each training round is budgeted 200 epochs for synthetic data and 1000 for food and haptic data. For all our experiments, the size of the overcomplete set ($S$) of informative triplets is twice the budget ($b$). We found that our method did not benefit from a larger $S$. This suggests it can be well suited to large datasets, though more testing is needed to confirm this.

\subsection{Batch and Initial Pool Sizes, Noise Impact, and Triplet Distance Measures (Q1)}

In general, as shown in Figure \ref{Q1}, our method achieves consistent improvements over the baselines. For all settings and datasets, multiple variants of our method outperform all baselines, often by large margins, though the same triplet distance measure is not always the most accurate. Relevant to practical settings, no variant of our method ever fails significantly: performance is maintained across the board.

\paragraph{Effect of batch size.} Intuitively, the benefits of decorrelation should be more significant for larger batches. Our experimental results confirm this hypothesis: as batch size increases, the performance gain of our method over random and uncertainty sampling is larger (Figure \ref{Q1}). Moreover, with increasing batch size, the performance of US also decreases, and sometimes becomes even worse than random. This further confirms that active learning based on just informativeness, without decorrelation, is ineffective.

\begin{figure}[!t]
\centering
\begin{tabular}{cc}
\hspace{-4.0mm} \includegraphics[scale=0.25]{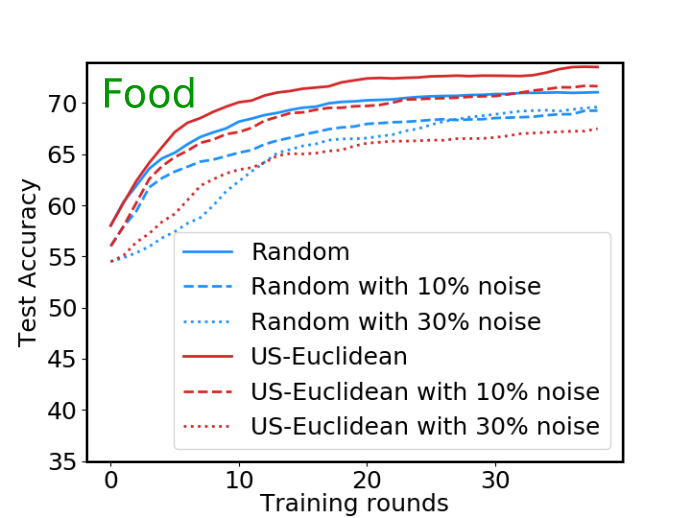}
\vspace{-2.0mm} \hspace{-5.75mm}
\includegraphics[scale=0.25]{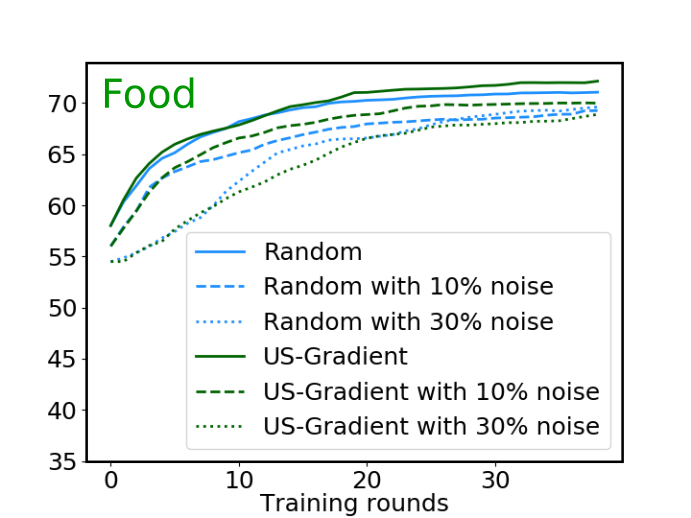}
\vspace{-0.0mm}
\end{tabular}
\caption{Performance of our method in the presence of additional labeling noise, for two different triplet decorrelation measures. Note that since this is a real-world dataset, some noise is likely present in the initial triplets as well.}
\label{noise}
\vspace{-3mm}
\end{figure}

\paragraph{Effect of initial pool.} Figure \ref{Q1} illustrates that an initial model trained on a very small pool of labeled data may not represent the underlying distribution. As a result, such a biased model may fail to sample informative triplets. As expected, our method outperforms baselines by larger margins when the initial pool is larger and more informative.

\paragraph{Effect of noisy labels.} To evaluate the noise robustness of our method, we induce additional noise in the data by randomly flipping the order of some triplets. In Figure \ref{noise}, we observe that the performance of all methods suffers from the increasing noise rate. However, our method outperforms random sampling, even in the presence of a moderate level of extra noise, and succumbs only to very high noise rates.

\paragraph{Effect of triplet distance measures.} Among different decorrelation measures, we found triplet gradient distance usually outperforms or matches variants and baselines. However, for very large batches, all variants of batch-decorrelated active learning perform better than standard baselines.

\subsection{Other Informativeness Measures (Q2)}

\begin{figure}[!t]
\centering
\begin{tabular}{cc}
\hspace{-4.0mm} \includegraphics[scale=0.25]{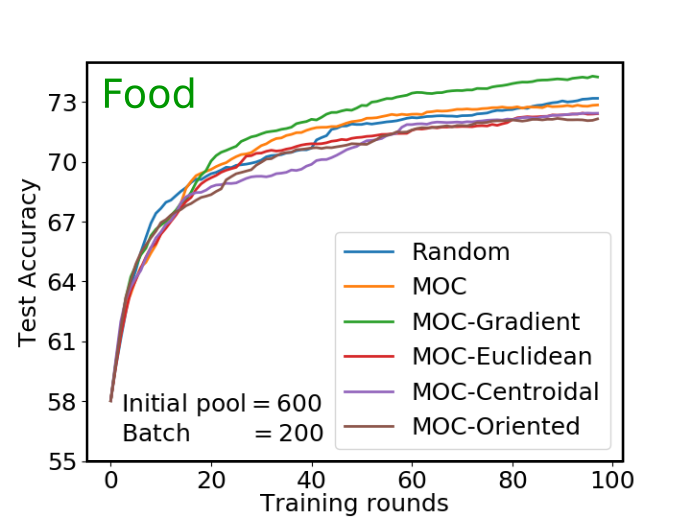}
\vspace{-3.0mm} \hspace{-5.75mm}
\includegraphics[scale=0.25]{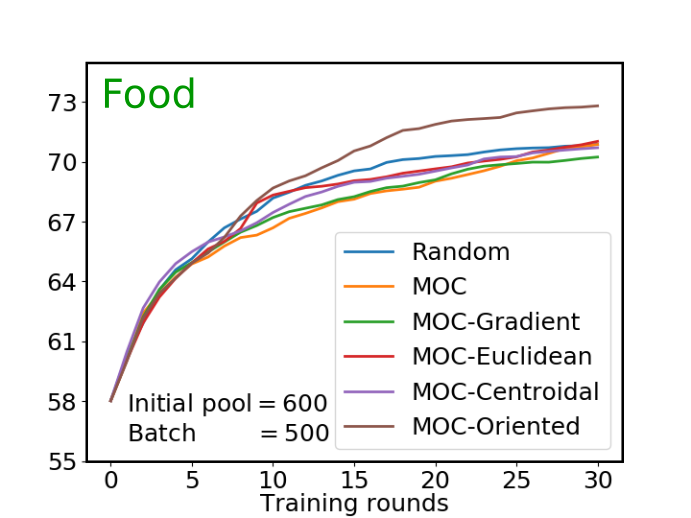}\\
\vspace{-3.0mm}
\hspace{-4.0mm}\includegraphics[scale=0.25]{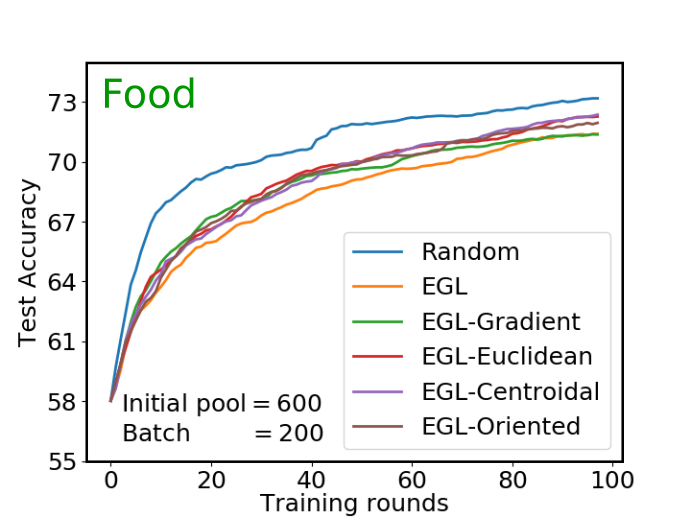}
\vspace{-3.0mm} \hspace{-5.75mm}
\includegraphics[scale=0.25]{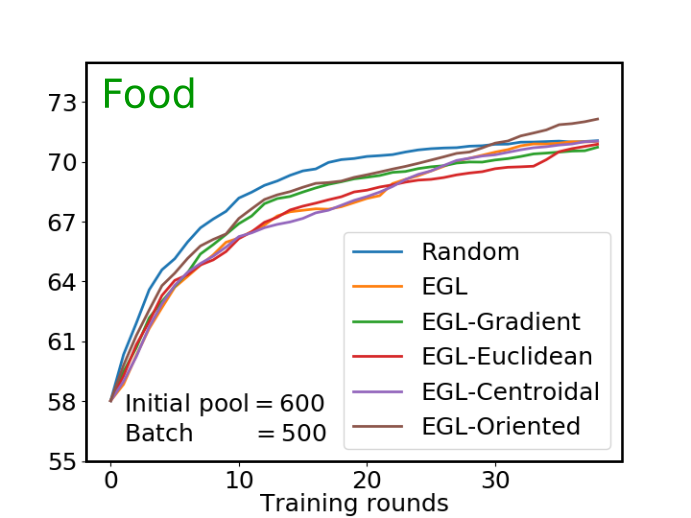}\\
\vspace{-1.0mm}
\end{tabular}
\caption{Performance of batch decorrelated AL with different triplet informativeness measures. EGL: Expected Gradient Length, MOC: Model Output Change.}
\label{haptic_var}
\vspace{-3mm}
\end{figure}

Since we decouple the choices for informativeness and diversity heuristics, we can easily examine other informativeness measures. Figure~\ref{haptic_var} shows the effect of replacing uncertainty (US) with (a) model output change (MOC) \cite{freytag2014selecting}, and (b) expected gradient length (EGL) \cite{huang2016active} on the Food dataset. We see that decorrelation improves both MOC and EGL, especially for larger batches, but does not always outperform random sampling, especially for EGL \cite{baldridge2009well}. This highlights the importance of an appropriate informativeness measure, and of decoupling the choice of this measure from triplet diversity.

\subsection{Ablation Studies (Q3)}

\begin{figure}[!t]
\centering
\vspace*{-1mm}
\includegraphics[scale=0.35]{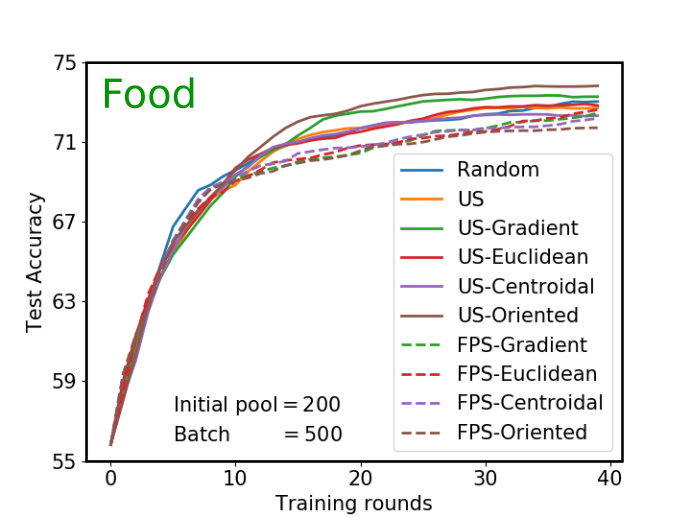}
\vspace*{-2mm}
\caption{Ablation studies showing the performance of individual components of our method.}
\label{ablation}
\vspace{-2mm}
\end{figure}

We compared triplet batch selection with the following variants: (1) just uncertainty-based informativeness ({\bf US}), (2) just diversity measure in different embedding spaces (\mbox{\bf FPS-{\em $\langle$Dist$\rangle$}}), and (3) combining both informativeness and diversity (\mbox{\bf US-{\em $\langle$Dist$\rangle$}}, i.e. our full method). As shown in Figure \ref{ablation}, active learning with combined measures performs much better than having informativeness or diversity alone. This again confirms the importance of diversifying a batch of informative objects for an effective active learning policy.

\section{Conclusion}

In this work, we presented a novel batch-mode active learning approach for triplet-based deep metric learning, that jointly balances the informativeness and diversity of a batch of triplets while decoupling the exact choice of heuristic for each criterion. We developed different measures for triplet decorrelation and found them effective for improving the performance of existing active learning approaches, which generally fail when labels are acquired in large batches. We further demonstrated that our decorrelation strategy generalizes easily to different informativeness measures. We extensively evaluated our method on three datasets and showed that our method is robust to different batch sizes, initial pools, and architectures, and performs consistently better than baselines.


\bibliographystyle{named}
\bibliography{ijcai20}

\end{document}